\documentclass[10pt,twocolumn,letterpaper]{article}

\usepackage{cvpr}
\usepackage{times}
\usepackage{epsfig}
\usepackage{graphicx}
\usepackage{tabularx}
\usepackage{threeparttable}
\usepackage{amsmath}
\usepackage{amssymb}
\usepackage[ruled,vlined]{algorithm2e}
\usepackage{color,soul}

\usepackage[pagebackref=false,breaklinks=true,letterpaper=true,colorlinks,bookmarks=false]{hyperref}

\cvprfinalcopy 

\def\cvprPaperID{***} 

\begin{document}

\title{ConceptLearner: Discovering Visual Concepts from Weakly Labeled Image Collections}

\author{Bolei Zhou$^\dag$, Vignesh Jagadeesh$^\ddag$, Robinson Piramuthu$^\ddag$\\
$^\dag$MIT $^\ddag$eBay Research Labs\\
{\tt\small bolei@mit.edu,[vjagadeesh, rpiramuthu]@ebay.com}}
\maketitle


\begin{abstract}


Discovering visual knowledge from weakly labeled data is crucial to scale up computer vision recognition system, since it is expensive to obtain fully labeled data for a large number of concept categories. 
In this paper, we propose ConceptLearner, which is a scalable approach to discover visual concepts from weakly labeled image collections. Thousands of visual concept detectors are learned automatically, without human in the loop for additional annotation. We show that these learned detectors could be applied to recognize concepts at image-level and to detect concepts at image region-level accurately. Under domain-specific supervision, we further evaluate the learned concepts for scene recognition on SUN database and for object detection on Pascal VOC 2007. ConceptLearner shows promising performance compared to fully supervised and weakly supervised methods.
\end{abstract}

\section{Introduction}
\label{sec:Introduction}
Recent advances in mobile devices, cloud storage and social network have increased the amount of visual data along with other auxiliary data such as text. Such big data is accumulating at an exponential rate and is typically diverse with a long tail. Detecting new concepts and trends automatically is vital to exploit the full potential of this data deluge. Scaling up visual recognition for such large data is an important topic in computer vision.  One of the challenges in scaling up visual recognition is to obtain fully labeled images for a large number of categories. The majority of data is not fully annotated. Often, they are mislabeled or labels are missing or annotations are not as precise as name-value pairs. It is almost impossible to annotate all the data with human in the loop. In computer vision research, there has been great effort to build large-scale fully labeled datasets by crowd sourcing, such as ImageNet~\cite{deng2009imagenet}, Pascal Visual Object Classes~\cite{pascal14}, Places Database~\cite{zhou2014nips} from which the state-of-the-art object/scene recognition and detection systems are trained~\cite{krizhevsky2012imagenet,girshick2013rich}. However, it is cumbersome and expensive to obtain such fully labeled datasets. Recently, there has been growing interest to harvest visual concepts from Internet search engines~\cite{chen2013neil,2014webly}. These approaches re-rank the search results and then learn concept detectors. The learned detectors largely depend on the quality of image search results, while image search engines themselves have sophisticated supervised training procedures. Alternatively, this paper explores another scalable direction to discover visual concepts from weakly labeled images.

\begin{figure}[t]
\begin{center}
\includegraphics[width=1\linewidth]{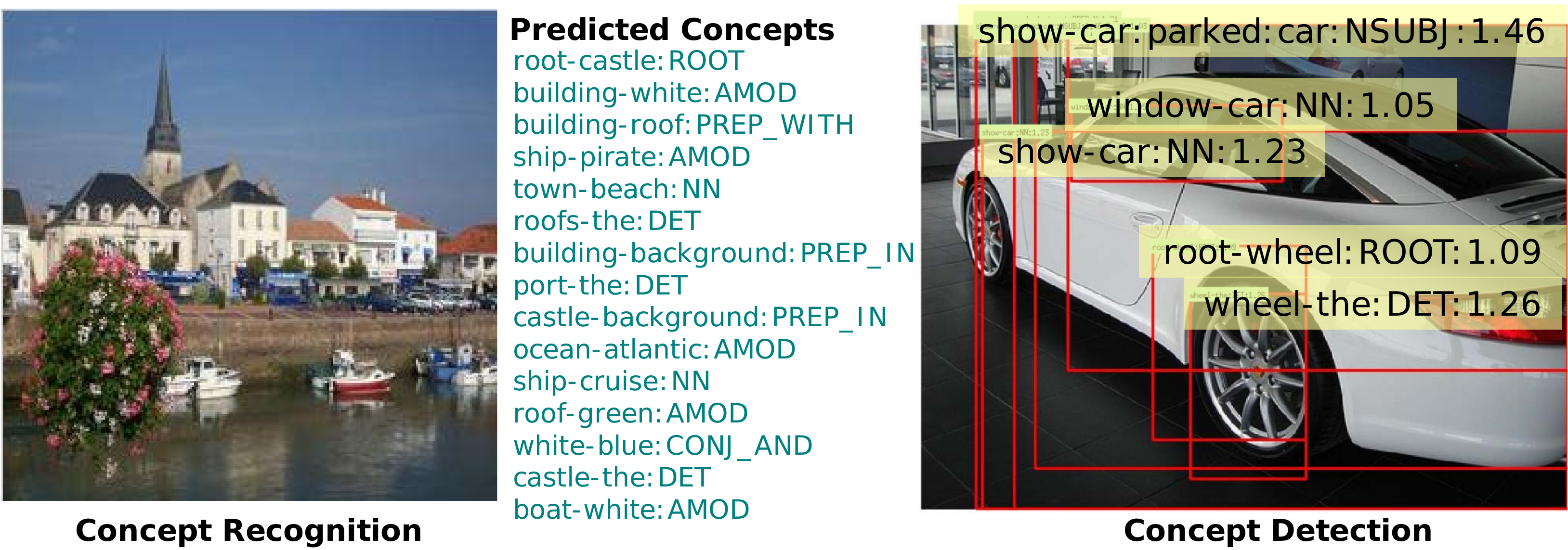}
\end{center}
 \caption{\textbf{ConceptLearner:} Thousands of visual concepts are learned automatically from weakly labeled image collections. Weak labels can be in the form of keywords or short description. ConceptLearner can be used to \emph{recognize} concepts at image level, as well as \emph{detect} concepts within an image. Here we show two examples done by the learned detectors.}
 \label{figure_cover}
\end{figure}

Weakly labeled images could be collected cheaply and massively. Images uploaded to photo sharing websites like Facebook, Flickr, Instagram typically include tags or sentence descriptions. These tags or descriptions, which might be relevant to the image contents, can be treated as weak labels for these images. Despite the noise in these weak labels, there is still a lot of useful information to describe the scene and objects in the image. Thus, discovering visual concepts from weakly labeled images is crucial and has wide applications such as large scale visual recognition, image retrieval, and scene understanding. Figure \ref{figure_cover} shows our concept recognition and detection results by detectors discovered by the ConceptLearner from weakly labeled image collections.

The contributions of this paper are as follows:
\begin{itemize}
\setlength{\itemsep}{0pt}
\setlength{\parskip}{0pt}
\item scalable max-margin algorithm to discover and learn visual concepts from weakly labeled image collections.
\item domain-selective supervision for application of weakly-learned concept classifiers on novel datasets.
\item application of learned visual concepts to the tasks of concept recognition and detection, with quantitative evaluation on scene recognition and object detection under the domain-selected supervision. 
\end{itemize}

The rest of the paper is organized as follows. Section~\ref{sec:RelatedWork} gives an overview of related work. Description of the model for weakly labeled image collections is in Section~\ref{sec:Modeling}. This is followed by max-margin visual concept discovery from weakly labeled image collections using hard instance learning in Section~\ref{sec:HardInstanceLearning}. Section~\ref{sec:DomainSpecific} shows how we can use the discovered concepts on a novel dataset using domain-selected supervision. We show 3 applications of concept discovery in Section~\ref{sec:Experiments}. We conclude with Section~\ref{sec:Conclusion} that gives a summary and a list of possible extensions. 


\section{Related Work}
\label{sec:RelatedWork}

Discovering visual knowledge without human annotation is a fascinating idea. Recently there have been a line of work on learning visual concepts and knowledge from  image search engines. For example, NEIL~\cite{chen2013neil} uses a semi-supervised learning algorithm to jointly discover common sense relationships and labels instances of the given visual categories; LEVAN~\cite{2014webly} harvests keywords from Google Ngram and uses them as structured queries to retrieve all the relevant diverse instances about one concept; \cite{li2013harvesting} proposes a multiple instance learning algorithm to learn mid-level visual concepts from image query results.

There are alternative approaches of discovering visual patterns from weakly labeled data that do not depend strongly on results from search engine. For example, \cite{berg2010automatic} uses multiple instance learning and boosting to discover attributes from images and associated textual description collected from the Internet. \cite{prest2012learning} learns object detectors from weakly annotated videos. \cite{wang2013weakly,song2014learning} use weakly supervised learning for object and attribute localization, where image-level labels are given and the goal is to localize these tags on image regions. \cite{singh2012unsupervised} learns discriminative patches as mid-level image descriptors without any text label associated with the learned patch patterns. In our work, we take on a more challenging task where both image and image-level labels are noisy in the weakly labeled image collections.

Other related work include \cite{ordonez2011im2text,kulkarni2011baby,gong2014improving,karpathy2014fragment}, which generate sentence description for images. They either generate sentences by image retrieval \cite{ordonez2011im2text}, or learn conditional random field among concepts \cite{kulkarni2011baby}, or utilize image-sentence embedding \cite{gong2014improving} and image-fragment embedding \cite{karpathy2014fragment} to generate sentences. Our work focuses more on learning general concept detectors from weakly labeled data. Note that the predicted labels obtained from our method could  also be used to generate sentence description, but it is beyond the scope of this paper.

%

\section{Modeling Weakly Labeled Images}
\label{sec:Modeling}
\begin{figure}[t]
\begin{center}
\includegraphics[width=1\linewidth]{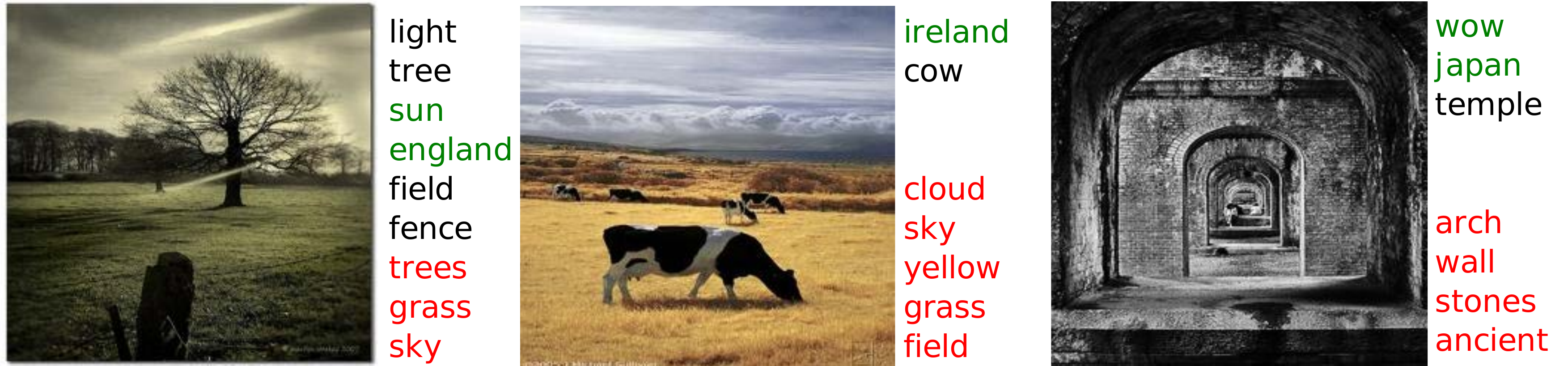}
\end{center}
 \caption{\textbf{NUS-WIDE Dataset~\cite{chua2009nus}:} Images have multiple tags/keywords. There are 1000 candidate tags in this dataset. Here are three examples, with original true tags shown in black, original noisy tags in green, and possible missing tags in red. }
\label{figure_NUSWIDE}
\end{figure}
\begin{figure}[t]
\begin{center}
\includegraphics[width=1\linewidth]{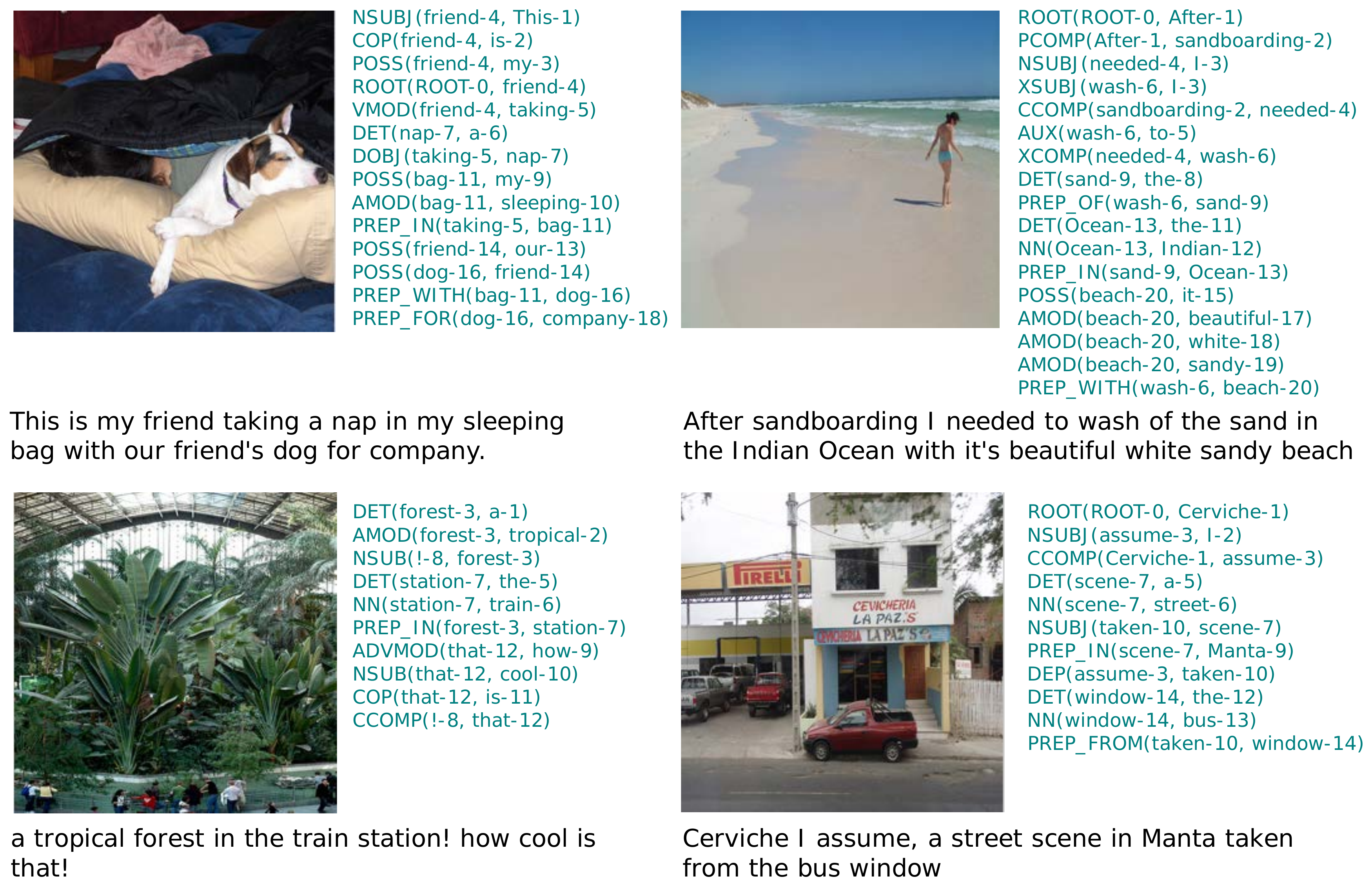}
\end{center}
 \caption{\textbf{SBU Dataset~\cite{ordonez2011im2text}:} Each image has a short description. Typically, this is a sentence, as shown below each image. We extract phrases from each sentence, as shown on the side of each image. Each phrase represents a relationship between two ordered words. The relationship is shown in capital letters. For example, AMOD dependency is like attribute+object, PREP are preposition phrases. Details of dependency types can be found in \cite{de2008stanford}.}
\label{figure_SBU}
\end{figure}


{\renewcommand{\arraystretch}{1.2}%
\begin{table}[t]
\small
\begin{threeparttable}[b]
\caption{Summary of notations used in this paper}
\label{table:Notation}
\begin{tabular}{| l | p{6.2cm} |}
\hline                       
\textbf{Variable}\tnote{1} & \textbf{Meaning}\\
\hline
$\mathcal{D}$   & Collection of weakly labeled images with associated tags  (which are used as weak labels)\newline(\ie) $\mathcal{D} = \{(I_i, \mathcal{T}_i) | I_i\in\mathcal{I}, \mathcal{T}_i\in\mathcal{T}\}_{i=1}^N$\\
$\mathcal{I}$   & Set of all images in $\mathcal{D}$\\
$\mathcal{T}$   & Set of unique tags in $\mathcal{D}$ (\ie)  $\bigcup_{i=1}^N\mathcal{T}_i$\\
$N$             & Number of images in $\mathcal{I}$ (\ie) $|\mathcal{I}|$\\
$T$             & Number of tags in $\mathcal{T}$ (\ie) $|\mathcal{T}|$\\
$I_i$           & An image in $\mathcal{I}$\\
$\mathcal{T}_i$ & The set of tags associated with $I_i$. For collections with sentence description for each image (as opposed to set of tags/keywords), the extracted phrases using \cite{de2008stanford} are the weak labels\\
$\tau_t$        & A tag in $\mathcal{T}$ (\ie) $\mathcal{T}=\{\tau_t\}_{t=1}^T$\\
$\mathcal{P}_t$ & Set of images associated with tag $\tau_t$ (\ie) $\mathcal{P}_t = \{I_i | \tau_t\in\mathcal{T}_i\}_{i=1}^{N}$\\
$\mathcal{N}_t$ & Set of images not associated with tag $\tau_t$ (\ie) $\mathcal{N}_t = \{I_i | \tau_t\not\in\mathcal{T}_i\}_{i=1}^{N}$\\
$V$ & Dimensionality of visual feature vector of an image\\
$\textbf{V}$ & Stacked visual features for $\mathcal{D}$. Row $i$ is a visual feature vector for image $I_i$.  $\textbf{V}\in\Re^{N\times V}$\\
$\textbf{T}$ & Stacked indicator vectors for $\mathcal{D}$. Row $t$ is an indicator vector for image $I_i$. Entry $(i,t)$ of $\textbf{T}$ is 1 when tag $\tau_t$ is associated with image $I_i$. It is 0 otherwise. $\textbf{T}\in[0,1]^{N\times T}$\\
$\textbf{w}_c$ & SVM weight vector, including the bias term for classifying concept $c$\\
$f_{\textbf{w}_c,\eta}^{hard}(\cdot)$ & Operator that takes a set of images and maps to \emph{hard} subset, based on SVM concept classifier $\textbf{w}_c$ such that $y\textbf{w}_c\cdot\textbf{x}<\eta$, where $\textbf{x}$ is the visual feature vector and $y\in\{-1,1\}$ label for  concept $c$\\
$f_{\textbf{w}_c,\eta}^{easy}(\cdot)$ & Operator, similar to $f_{\textbf{w}_c,\eta}^{hard}(\cdot)$, that takes a set of images and maps to \emph{easy} subset, such that $y\textbf{w}_c\cdot\textbf{x}>\eta$\\
$Rand_k(\cdot)$ & Operator that takes a set of images and randomly pick $k$ images without replacement. (\ie) 
$Rand_k(\mathcal{I}_{s}) =\{I_{r(j)} | I_{r(j)}\in\mathcal{I}_{s}, \mathcal{I}_{s}\subset\mathcal{I}\}_{j=1}^k$, where $r(j)$ picks a unique random integer from $\{i | I_i\in\mathcal{I}_{s}\}$\\
\hline  
\end{tabular}
\begin{tablenotes}
\item[1] Sets are denoted by scripts, matrices by bold upper case, vectors by bold lower case, scalars by normal faced lower or upper case.
\end{tablenotes}
\end{threeparttable}
\vspace{-3em}
\end{table}
}
\addtocounter{footnote}{1}

\begin{algorithm}[t]
\caption{ConceptLearner}
\label{discoveryAlgorithm}
\KwData
{ 
See Table~\ref{table:Notation} for notations.\\
\hspace{1cm}(i) $\textbf{V}$, matrix of visual feature vectors \\
\hspace{1cm}(ii) $\textbf{T}$, matrix of tag indicator vectors 
\begin{description}
\item \hspace{-0.5em}\textbf{Parameters:}\\
(i) $\alpha$, ratio of cardinalities of negative and positive instance sets\\
(ii) $M_{t}$, number of image clusters for tag $\tau_t$ \\
(iii) $\eta$, threshold to determine hard and easy instances\\
(iv) $K$, the top number of tags based on tf-idf for each concept cluster.
\end{description}
}

\KwResult
{
(i) Matrix $\textbf{W}$ of SVM weight vectors, where $c^{th}$ row is concept detector $\textbf{w}_{c}^T$ 
(ii) name set for each concept $c$
}

\For{label $t=1:T$}
{
	$c = 0;$ \emph{/* Initialize concept count */}\\
    
    Construct $\mathcal{P}_t$, $\mathcal{N}_t$\;
    
    Use $\textbf{V}, \textbf{T}$ to cluster images $\mathcal{P}_t$ into $M_{t}$ clusters. \\
    
    Each such cluster is a concept\;
    
    \For{cluster $m=1:M_{t}$}
    {
    	$c = c + 1$\;
        
        Construct the positive training set $\mathcal{P}_t^{train}:=\{I_i | I_i\in\mathcal{P}_t, I_i\in \mbox{cluster} \ m\}$\;
        
        $N_p := \left|\mathcal{P}_t^{train}\right|$, size of positive training set\;
        
        $N_n := \lceil\alpha N_p\rceil$, size of negative training set\;
        
        Initialize the negative training set $\mathcal{N}_t^{train} \leftarrow Rand_{N_n}\left(\mathcal{N}_t\right)$\;
        
\emph{/* Fix $\mathcal{P}_t^{train}$ and mine hard negative instances */}\\

	    \While{$\mathcal{N}_t^{train}$ is updated}
        {
          Train SVM on $\mathcal{P}_t^{train}$ and $\mathcal{N}_t^{train}$ to get weight vector $\textbf{w}_{c}$\;

  Easy positives $\mathcal{P}_t^{easy} := f_{\textbf{w}_c,\eta}^{easy}(\mathcal{P}_t^{train})$\;
  
  Hard negatives $\mathcal{N}_t^{hard} := f_{\textbf{w}_c,\eta}^{hard}(\mathcal{N}_t^{train})$\;

  Easy negatives $\mathcal{N}_t^{easy} := f_{\textbf{w}_c,\eta}^{easy}(\mathcal{N}_t^{train})$\;

  Update $\mathcal{N}_t^{train} \leftarrow \mathcal{N}_t^{hard} \bigcup Rand_{N_n-\left|\mathcal{N}_t^{easy}\right|}\left(\mathcal{N}_t	\setminus \mathcal{N}_t^{easy}\right)$\;
  }

  \emph{/* Cache tag frequency for the positive set */}\\
  Calculate tag frequency vector $\textbf{f}_{m}\in\mathbb{Z}_{\ge 0}^{T}$ based on images in $\mathcal{P}_t^{easy}$\;    
      }
      
\emph{/* Name each concept using tf-idf across the label frequencies, w.r.t. $M_t$ clusters */}\\
Compute the tf-idf based on $\{\textbf{f}_{1},\textbf{f}_{2},...,\textbf{f}_{M_t}\}$\;
    
Create a name set for each concept $m\in[1,M_t]$, by taking the top $K$ labels based on tf-idf\;
}
\end{algorithm}


Generally speaking, there are two categories of weakly labeled image collections: (i) multiple tags for each image as in NUS-WIDE dataset~\cite{chua2009nus} and (ii) sentence description for each image as in SBU dataset~\cite{ordonez2011im2text}. Here we analyze the representative weakly labeled image collections NUS-WIDE and SBU dataset respectively. 


Figures~\ref{figure_NUSWIDE} and~\ref{figure_SBU} illustrate samples from NUS-WIDE dataset~\cite{chua2009nus} and SBU dataset~\cite{ordonez2011im2text}. Note that tags in Figure~\ref{figure_NUSWIDE} can be incorrect or missing. In Figure~\ref{figure_SBU} sentences associated with images in ~\cite{ordonez2011im2text} are also noisy, as they were written by the image owners when the images were uploaded. Image owners usually selectively describe the image content with personal feelings, beyond the image content itself. 

There is another category of image collection with sentence description such as Pascal Sentence dataset~\cite{rashtchian2010collecting} and Pascal30K dataset~\cite{hodoshimage}. These sentence descriptions are generated by the paid Amazon Mechanical Turk workers rather than the image owners, and are more objective and accurate to the image contents. However the labeling is expensive and not scalable to millions of images. Our approach could work on all of the three categories of weakly labeled image collections, but we focus on the first two more challenging categories.

For image collections with multiple tags, we just take the sparse tag count vector as the weak label feature of each image. For image collections with sentence description, we extract phrases, which are semantic fragments of sentence, as weak label feature for each image. A sentence contains not only several entities such as multiple weak tags for the image, but also contains relationships between the entities. These relationships between entities, composed as phrases, could be easily interpreted and effectively used by human. The phrase representation is more descriptive than a single keyword to describe the image content.  Figure~\ref{figure_SBU} shows some examples of extracted phrases from  sentences. For simplicity, we adopt the Stanford typed dependencies system \cite{de2008stanford} as the standard for sentence parsing. All sentences are parsed into short phrases and only those that occur more than  50 times are kept. Note that in \cite{sadeghi2011recognition}, 17 visual phrases are manually defined and labeled, corresponding to chunks of meaning bigger than objects and smaller than scenes as intermediate descriptor of the image.  In contrast, our approach is data-driven and extracts thousands of phrases from image sentence descriptions automatically. We use these extracted phrases as weak labels for images and learn visual concepts automatically at scale.

Notations used in this paper are summarized in Table~\ref{table:Notation}.



\section{Max Margin Visual Concept Discovery}
\label{sec:HardInstanceLearning}

{\renewcommand{\arraystretch}{1.2}%
\begin{figure}[t]
\centering
\small
\begin{tabular}{p{3.5cm} p{4cm}}
\hline
flower-yellow:AMOD       & view-tower:PREP\_FROM \\
boats-house:NN           & clouds-sky:PREP\_AGAINST\\
table-chair:CONJ\_AND    & walking-beach:PREP\_ALONG \\
flowers-field:PREP\_IN   & bridge-lake:PREP\_OVER\\
sand-beach:PREP\_AT      & canoe-river:PREP\_DOWN\\
waiting-train:PREP\_FOR  & grass-sky:PREP\_AGAINST\\
\hline\\ [-2ex]
\end{tabular}
\caption{\textbf{Sentence to Phrases:} Example phrases extracted from the sentences of SBU dataset. Each phrase shows a pair of words and the relationship between them. See~\cite{de2008stanford} for details of the relationship. We use each phrase to represent a concept and group the associated images together. Each group is then refined using Algorithm~\ref{discoveryAlgorithm}. This refined collection of groups is then used to learn concept classifiers and detectors.}
\label{figure_phraseText}
\end{figure}
}


Learning visual patterns from weakly labeled image collection is challenging because the labels for training images are noisy. Existing learning methods for this task include semi-supervised learning as in \cite{chen2013neil} and multiple instance learning as in \cite{berg2010automatic,li2013harvesting}. In this paper, we formulate this problem as max-margin hard instance learning of visual concepts using SVM.

Since the labels for every image are noisy and there are a lot of missing labels, there is no clear separation of positive set and negative set. If the images with a specific label are considered as the positive images for that label and images without that label as the negative images, there would be a lot of false positives (image with some concept label but has no noticeable image content related to that concept) in the positive set and false negatives (image with some visible concept inside but without that concept labeled) in the negative set. Inspired by the idea of hard instance mining used in face detection and object detection \cite{dalal2005histograms,felzenszwalb2010object}, we consider false positives and false negatives as hard instances in the learning of visual concepts. The algorithm will iteratively seek the max-margin decision boundary that separates hard instances.


The detailed steps of our algorithm for concept discovery are listed in Algorithm~\ref{discoveryAlgorithm}. Our algorithm starts with an initial cache of instances, where the positive set includes all the examples with label $t$ and the negative set is a random sample of images without that label $t$. In each iteration, we remove easy instances from the cache and add additional randomly selected negative images. The SVM is then re-trained on the new cache of positive and negative sets. Here we keep the positive set fixed and only do hard negative instance sampling.

$\alpha$ is the ratio of the size of negatives over the size of positives.  Since the number of hard negative instance might be high, we keep a relatively large ratio  $\alpha = 5\sim10$. On the other hand, as there are various views or sub-categories related to the same concept, it is better to learn several sub-category detectors for the same concept than to learn a single detector using all the positive set. Thus we do clustering on the positive sets before learning concept detectors. The cluster number $M_{t}$ for $t^{th}$ tag controls the diversity of the learned detectors. 
 Tf–idf~\cite{tfidf}, short for term frequency inverse document frequency, is used to find the important contextual labels in the label frequency for each sub-categories so that we could better name each learned sub-category detectors.

\section{Selecting Domain-Specific Detectors}
\label{sec:DomainSpecific}
After the concept detectors are learned, we could directly apply all of them for concept recognition at image-level. But in some applications, we need to apply one concept detector or subset of detectors from the pool of detectors learned from source dataset (say, SBU) to some specific tasks on target dataset (say, Pascal VOC 2007). Here we simply use a winner-take-all selection protocol for the detector selection. We define a selection set, 
which contains some labeled instances from the target dataset. Then the relevant concept detector with the highest accuracy/precision on the target dataset is selected. Note that the selection set should be separated from the test set of the target dataset. In the following experiments on scene recognition and object detection, we follow this selection protocol to automatically select the most relevant detectors for evaluation on test set. We call this as \textit{domain selected supervision}. This is related to the topic of domain adaptation \cite{saha2011active,tang2012shifting}, but we do not use the instances in the target domain to fine-tune the learned detectors. Instead, we only use a small subset of the target domain to select the most relevant concept detectors from a large pool of pre-trained concept detectors. It is also related to the issue of dataset bias \cite{torralba2011unbiased} existing in current recognition datasets. Domain-selected supervision provides a nice way to generalize the learned detectors to novel datasets.

\section{Experiments}
\label{sec:Experiments}


We evaluate the learning of visual concepts on two weakly labeled image collections: NUS-WIDE~\cite{chua2009nus} and SBU~\cite{ordonez2011im2text} datasets. NUS-WIDE has 226,484 images (the original set has 269,649 URLs but some fo them are invalid now) with 1000 tags (which were used as weak labels) and 81 ground-truth labels. As shown in \cite{chua2009nus}, the average precision and recall of tags with the corresponding ground-truth labels are both about 0.5, which indicates that about half of the tags are incorrect and half of the true labels are missing. We acquired 934,987 images (the original set has 1M URLs but some of them are invalid now) from SBU dataset. Each image has a text description written by the image owner. Examples from these two datasets are shown in Figures~\ref{figure_NUSWIDE} and~\ref{figure_SBU}.

The 4096 dimensional feature vector from the FC7 layer of Caffe reference network~\cite{Jia13caffe} was used as the visual feature for each image, since deep features from pre-trained Convolutional Neural Network on ImageNet~\cite{imagenet09} has shown state-of-the-art performance on various visual recognition tasks~\cite{razavian2014cnn}. Each description was converted to phrases using the Stanford English Parser~\cite{de2006generating}. Phrases with count smaller than 50 were not used. We used 7437 phrases. Figure~\ref{figure_phraseText} shows some sample phrases. We could see that these phrases contain rich information, such as relationships attribute-object, object-scene, and object-object.  We use linear SVM from liblinear~\cite{fan2008liblinear} in the concept discovery algorithm. 


\begin{figure}[t]
\begin{center}
\includegraphics[width=1\linewidth]{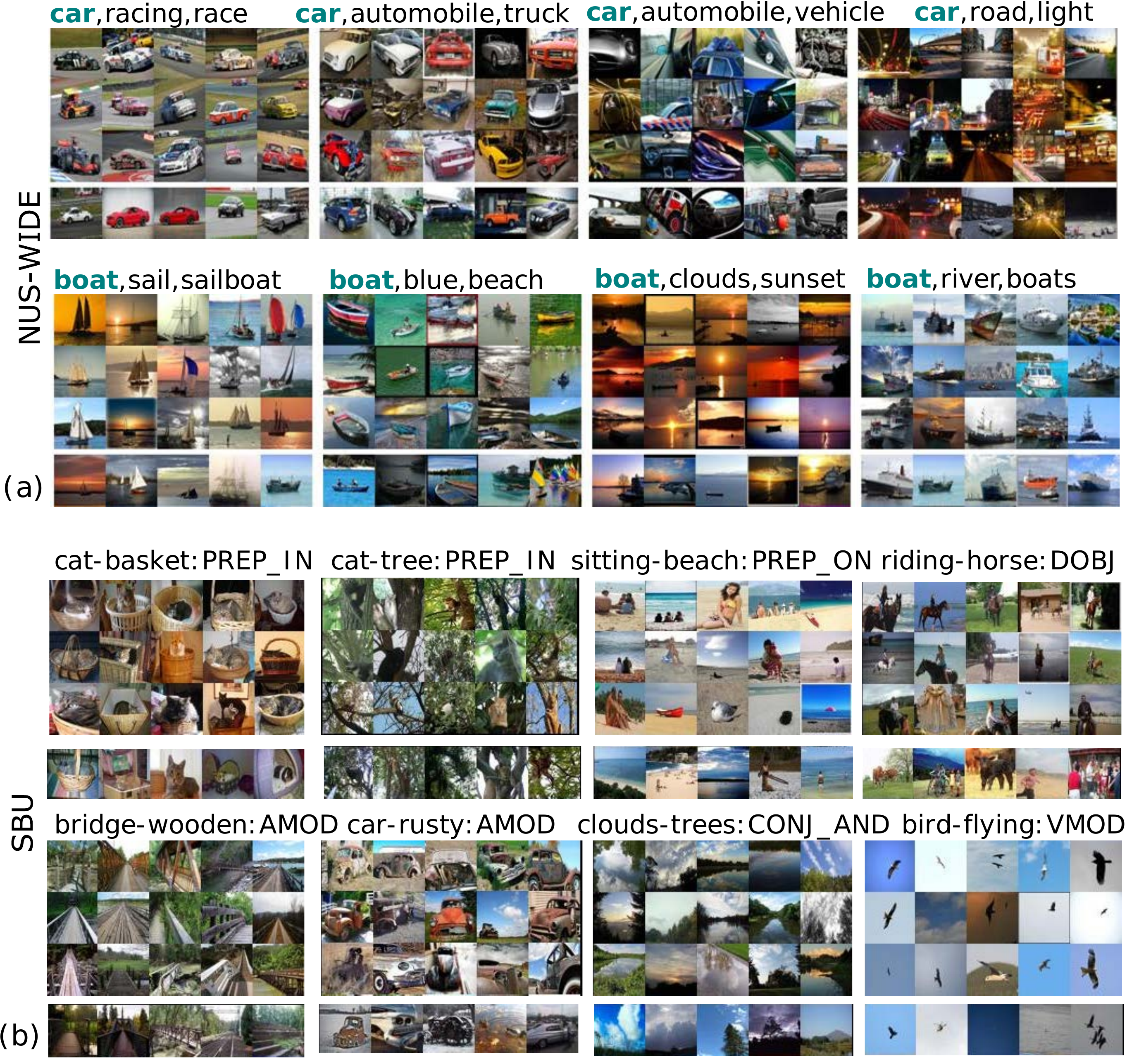}
\end{center}
 \caption{\textbf{Discovered Concepts:}  Illustration of the learned concepts from NUS-WIDE and SBU datasets. Each montage contains the top 15 positive images for each concept, followed by a single row of negative images. 4 sub-category concept detectors for car and boat respectively are illustrated in (a), based on concepts learned from NUS-WIDE. The title shows the name set for each concept from NUS-WIDE. Phrases for SBU dataset are shown in titles as in (b).}
\label{figure_learnedDetector}
\end{figure}

Concepts were learned independently from these datasets using Algorithm~\ref{discoveryAlgorithm}. Once concepts were learned, we consider 3 different applications: (i) concept detection and recognition, (ii) scene recognition and (iii) object detection. For concept detection and recognition, we chose $M_{t}=1$ and $M_{t}=4$ for learning concepts from SBU and NUS-WIDE datasets respectively. For scene recognition and object detection, we varied $M_{t}=1\sim 10$ to learn the selected concepts and then pooled together all  possible concept detectors. Note that $M_{t}$ was determined empirically, a larger $M_{t}$ might generate near-duplicate or redundant concept detectors, but it might make the concept pool more diverse. Determining $M_{t}$ automatically for each label $t$ is part of future work. The illustration of some learned concept detectors along with the top ranked positive images is shown in Figure~\ref{figure_learnedDetector}. 

For the concepts learned from NUS-WIDE dataset in Figure~\ref{figure_learnedDetector}(a), we show the central concept (cat, boat) in each row along with their variations. The title show 3 tags of which the first one is the central concept. The other two tags are more contextual words ranked from tf-idf scores associated with the central concept name as the sub-category concept name. We can see that there are indeed sub-categories representing different views of the same concepts, the contextual words ranked using tf-idf well describe the diversity of the same concept. For the concepts learned from SBU dataset, we show 8 learned phrase detectors in Figure~\ref{figure_learnedDetector}(b). We can see that the visual concepts well match the associated phrases. For example, cat-in-basket and cat-in-tree describe the cat in different scene contexts; sitting-on-beach and riding-horse describe the specific actions; wooden-bridge and rusty-car describe the attributes of objects. Besides, the top ranked hard negatives are also shown below the ranked positive images. We can see that these hard negatives are visually similar to the images in the positive set.

To evaluate the learned concept detectors, we use images from the SUN database~\cite{xiao2010sun} and Pascal VOC 2007 object detection dataset~\cite{pascal14}. These are independent from the NUS-WIDE and SBU datasets where we discover the concept detectors. We first show some qualitative results of concept recognition and detection done by the learned detectors. Then we perform quantitative experiments to evaluate the learned concept detectors on specific vision tasks through domain-selected supervision, for scene recognition and object detection respectively. Compared to the fully supervised methods and weakly supervised methods, our domain-selected detectors show very promising performance\footnote{More experimental results are included in supplementary materials}. 

\begin{figure*}[t]
\begin{center}
\includegraphics[width=1\linewidth]{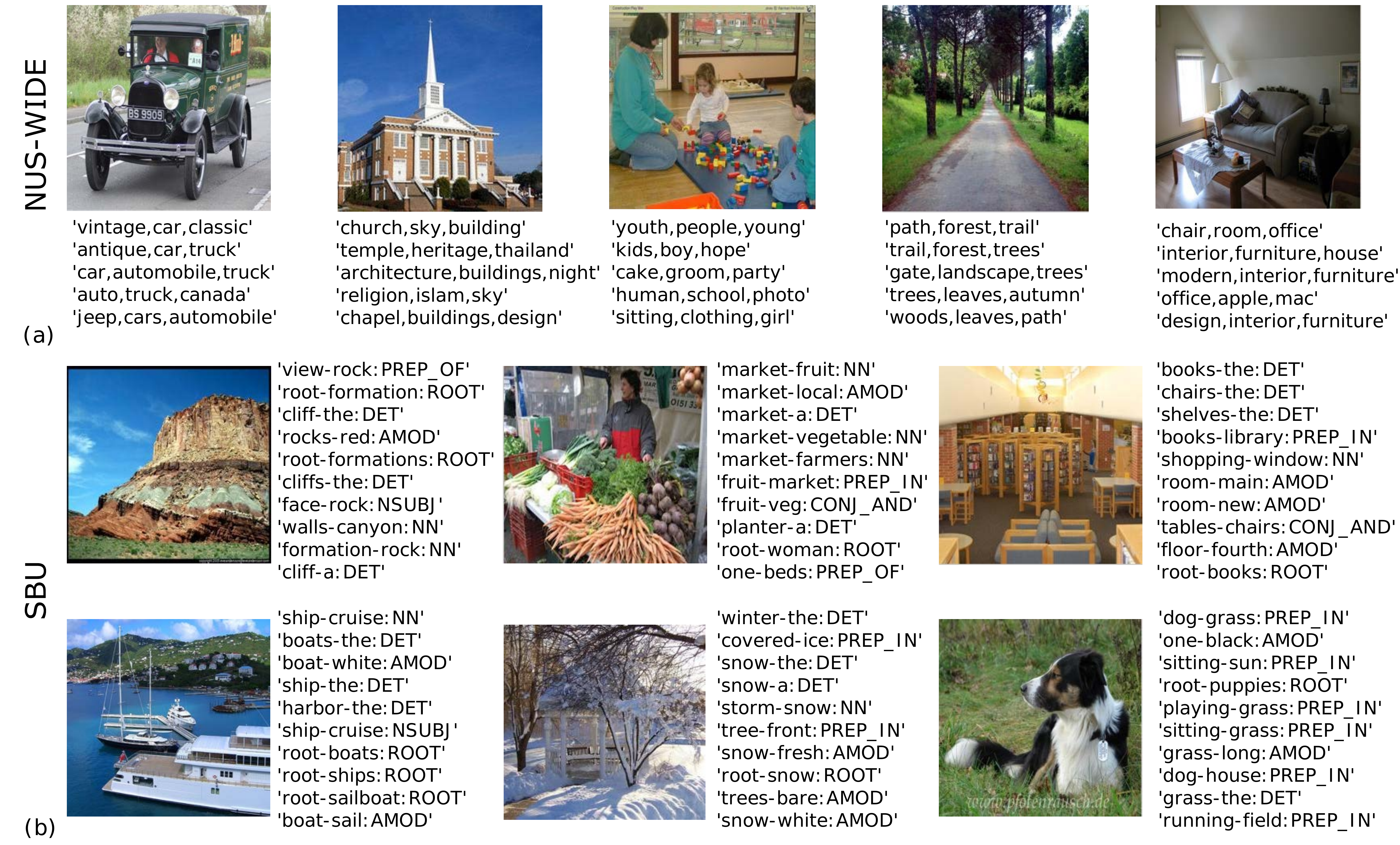}
\end{center}
 \caption{\textbf{Concept Recognition:} Illustration of concept recognition using concepts discovered from  (a) NUS-WIDE and (b) SBU datasets. Top 5 and 15 ranked concepts are shown respectively.  These predicted concepts well describe the objects, the scene contexts, and the activities in these images.}
\label{figure_imagelevelannotation}
\end{figure*}

\subsection{Concept Recognition and Detection}
\label{subsec:Concept}
We apply the learned concept detectors for concept recognition at image level and concept detection at image regions. After the deep feature $\textbf{x}_q$ for a novel query image $I_q$ is extracted, we multiply the learned detector matrix with the feature vector to get the response vector $\textbf{r} = \textbf{W}\textbf{x}_q$, where each element of the vector is the response value of one concept. Then we pick the most likely concepts of that image by simply sorting the response values on $\textbf{r}$. 

We randomly take the images from SUN database \cite{xiao2010sun} and Pascal VOC 2007 as query images, the recognition results by concept detectors learned from NUS-WIDE and SBU datasets are shown in Figure~\ref{figure_imagelevelannotation}. We can see that the predicted concepts well describe the image contents, from various aspects of description, such as attributes, objects and scenes, and activities in the image. 

\begin{figure}[h]
\begin{center}
\includegraphics[width=1\linewidth]{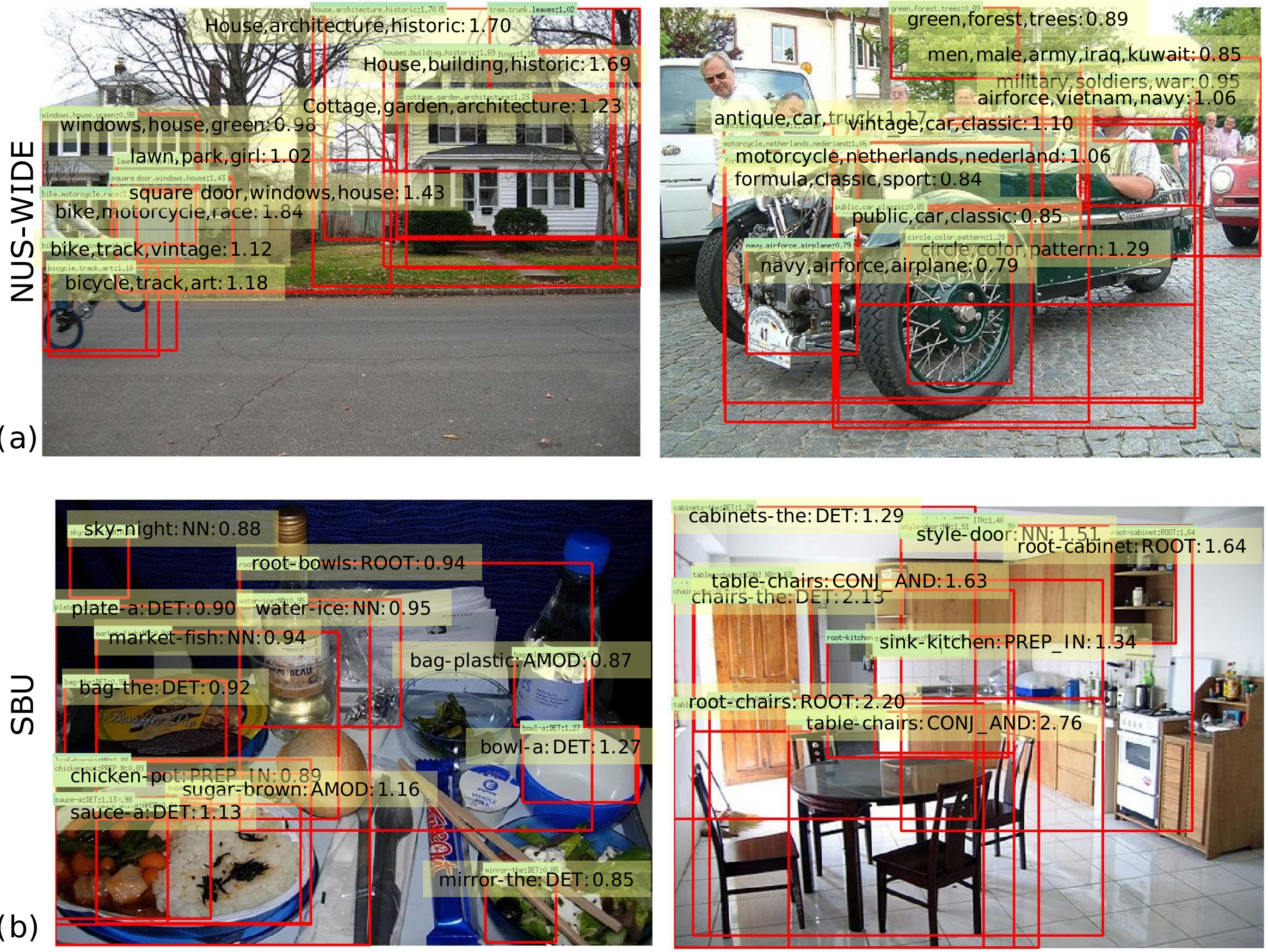}
\end{center}
 \caption{\textbf{Concept detection:} Results of concepts discovered from (a) NUS-WIDE and (b) SBU. Top 20 bounding boxes with high detector responses are shown.  Note that for legibility we manually overlaid the text labels with large fonts.}
\label{figure_detection}
\vspace{-1ex}
\end{figure}


\begin{figure*}[t]
\begin{center}
\includegraphics[width=1\linewidth]{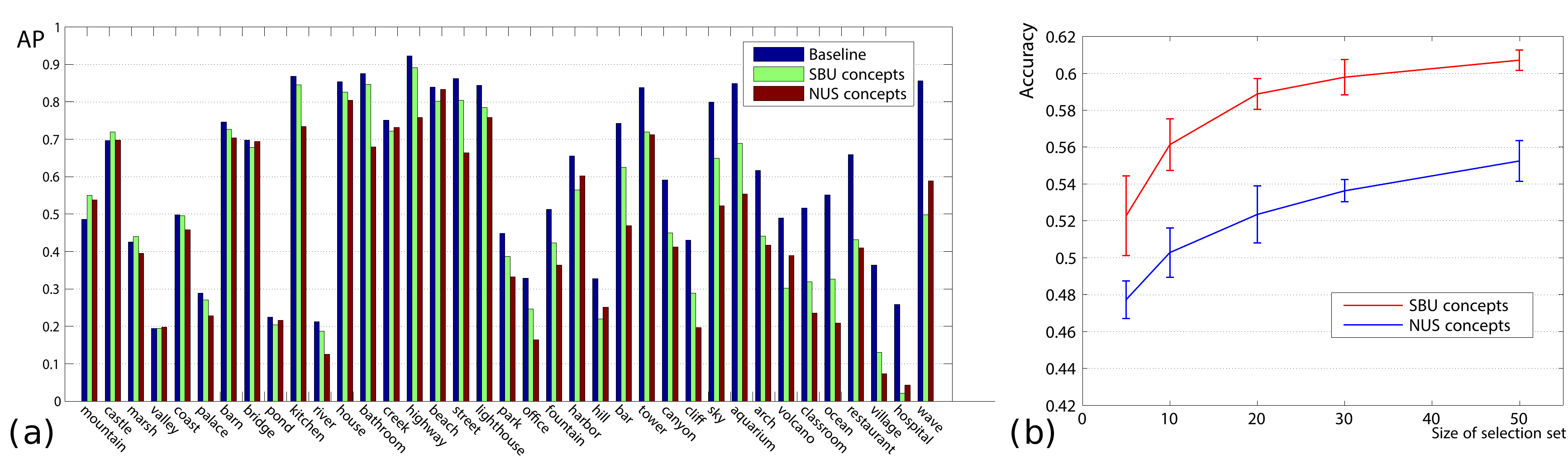}
\end{center}
\vspace{-1em}
 \caption{\textbf{Scene Recognition on SUN:} (a) AP per category for three methods, ranked by the gap between the learned concepts and fully supervised baseline. SBU concept detectors from weak labels outperform the baseline for mountain, castle, marsh, and valley. The concept detectors perform worse for village, hospital, and wave, due to the lack of sufficient positive examples in the weakly labeled image collections (b) Recognition accuracy over the size of selection set. Domain-specific detectors work well when there are only a few samples in the selection set.}
\vspace{-1em}
\label{figure_confusion}
\end{figure*}

Furthermore, we could apply the learned concept detectors for concept detection at the level of image regions. Specifically, we mount the learned concept detectors on a detection system similar to the front-end of Region-CNN~\cite{girshick2013rich}: Selective search~\cite{uijlings2013selective} is first used to extract region proposals from the test image. Then CNN features of region proposals are extracted. Finally the deep features of every region proposals are multiplied with the detector matrix and non-maximum suppression is used to merge the responses of the overlapped region proposals. The concept detection results are shown in Figure~\ref{figure_detection}. We can see that this simple detection system mounted with learned concept detectors interprets the images in great detail.

\subsection{Scene Recognition on SUN database}
\label{subsec:SUN}
Here we evaluate the learned concept detectors for the scene recognition on the SUN database \cite{xiao2010sun} which has 397 scene categories. We firstly use the scene name to select the relevant concept detectors from the pool of learned concepts \ie the scene name appears in the name of some concept detector. There are 37 matched scene categories among the concept pool of SBU and the concept pool of NUS-WIDE. We take all the images of these 37 scene categories from SUN database and randomly split them into train and test sets. The size of train set is 50 images per category. We train a linear SVM on the train set as the fully supervised baseline. Note that this baseline is quite strong, since linear SVM plus deep feature is currently the state-of-the-art single feature classifier on the SUN database~\cite{zhou2014nips}.  

{\renewcommand{\arraystretch}{1.2}%
\begin{table}[t]
\small
\caption{Accuracy and mean average precision (mAP) of baseline, NUS-WIDE concepts and SBU concepts. Mean $\pm$ std is computed from 5 random splits of training and testing. }
\label{table:sceneclassification}
\begin{tabular}{ |l| l | c  |c|}
 \hline                       
  Method & Supervision & Accuracy & mAP \\
\hline
Baseline (strong) & Full & 69.0$\pm$0.6 & 59.6$\pm$0.8 \\
NUS concepts (weak)& Selected & 55.5$\pm$1.8& 47.0$\pm$0.4\\
SBU concepts (weak)& Selected & 60.0$\pm$1.2& 50.6$\pm$0.7\\
\hline  
\end{tabular}
\end{table}
}

To evaluate the learned concepts, we use the domain-selected supervision introduced in Section~\ref{sec:DomainSpecific}. The train set is used as the selection set. 37 best scene detectors are selected out from the concept pool of SBU and NUS-WIDE based on their top mAP on the selection set, then they are evaluated on the test set. A test image is classified into the scene category which has the highest detector response. Without the calibration of the detector responses, the classification result is already reasonsably good. 

The accuracy and mean average precision (mAP) of the fully supervised baseline and our domain-selected supervised methods are listed in Table \ref{table:sceneclassification}. The AP per category for the three methods are plotted in Figure~\ref{figure_confusion}(a). We can see that the SBU concept detectors perform better than the NUS-WIDE concept detectors because of larger amount of data. Both of the learned concept detectors have good performance, compared to the fully supervised baseline with strong labels. SBU concept detectors even outperform the baseline for mountain, castle, marsh, and valley categories shown in Figure~\ref{figure_confusion}(a). The concept detectors perform worse on some scene categories like village, hospital, and wave, because there are not so many good positive examples in the weakly labeled image collections.

In Figure \ref{figure_confusion}(b), we further analyze the influence of selection set on the performance of our method. We randomly select the subset of images from the train set as the selection set for our method, we can see that the SBU concepts still achieve 52.5\% accuracy when there are only 5 instances per category as the selection set to pick the most relevant concept detectors. It shows that the domain-selected supervision works well even with few samples from the target domain.


\subsection{Object Detection on Pascal VOC 2007}
\label{subsec:Pascal}

{\renewcommand{\arraystretch}{1.2}%
\begin{table*}[!ht]
 \caption{Comparison of methods with various kinds of supervision on Pascal VOC 2007. NUS-WIDE has missing entries since some object classes don't appear in the original tags.}
 \label{figure_detectionTable}
\scriptsize
\noindent\begin{tabular}{ |p{1.3cm} | p{1cm} | p{0.25cm}  p{0.25cm}  p{0.25cm}  p{0.25cm} p{0.25cm} p{0.25cm} p{0.25cm} p{0.25cm} p{0.25cm} p{0.25cm} p{0.25cm} p{0.25cm} p{0.25cm} p{0.25cm} p{0.25cm} p{0.25cm} p{0.25cm} p{0.25cm} p{0.25cm} p{0.3cm} |p{0.35cm}|}
 \hline                       
  Method & Supervision & aero & bike & bird & boat & bottle & bus & car & cat & chair & cow & table & dog & horse & mbik & pers & plant & sheep & sofa & train & tv & mAP\\
\hline
SBU & Selected & 34.5 & \textbf{39.0} & \textbf{18.2}& 14.8& 8.4& 31.0& 39.1& 20.4& \textbf{15.5}& 13.1& \textbf{14.5}& 3.6& 20.6& 33.9& \textbf{9.4}& \textbf{17.0}& 14.7& \textbf{22.6}& 27.9& \textbf{19.0} & \textbf{20.9}\\
  NUS-WIDE & Selected & \textbf{34.6} & 38.5 & 16.5& \textbf{18.7}& -& 27.0& \textbf{43.6}& \textbf{24.6}& 10.9& 9.3& -& \textbf{20.4}& 30.3& \textbf{36.6}& 3.0& 4.7& 13.6& -& \textbf{36.1}& - &-\\
   
  CVPR'14~\cite{2014webly} & Webly & 14.0&36.2&12.5&10.3&\textbf{9.2}&\textbf{35.0}&35.9&8.4&10.0&\textbf{17.5}&6.5&12.9&\textbf{30.6}&27.5&6.0&1.5&\textbf{18.8}&10.3&23.5&16.4&17.2\\
   ECCV'12~\cite{prest2012learning} & Video & 17.4&-&9.3&9.2&-&-&35.7&9.4&-&9.7&-&3.3&16.2&27.3&-&-&-&-&15.0&-&- \\
\hline  
   ICCV'11~\cite{siva2011weakly} & Weakly & 13.4 & 44.0 &3.1& 3.1& 0.0& 31.2& 43.9& 7.1& 0.1& 9.3& 9.9& 1.5& 29.4& 38.3& 4.6& 0.1& 0.4& 3.8& 34.2& 0.0 &13.9\\
  ICML'14~\cite{song2014learning} & Weakly & 7.6& 41.9& 19.7& 9.1& 10.4& 35.8& 39.1& 33.6& 0.6& 20.9& 10.0& 27.7& 29.4& 39.2& 9.1& 19.3& 20.5& 17.1& 35.6& 7.1&22.7\\
  CVPR'14~\cite{girshick2013rich} & Full & 57.6 &57.9& 38.5& 31.8& 23.7& 51.2& 58.9& 51.4& 20.0& 50.5& 40.9& 46.0& 51.6& 55.9& 43.3& 23.3& 48.1& 35.3& 51.0& 57.4 &44.7\\
\hline  
\end{tabular}
\vspace{-2em}
\end{table*}
}
We further evaluate the concept detectors on Pascal VOC 2007 object detection dataset. We follow the pipeline of the region proposal and deep feature extraction in~\cite{girshick2013rich} for the validation and test sets of Pascal VOC 2007. Under domain-selected supervision, we first select the learned concept detectors which have the object name inside their name and compute the AP for each of them on the validation set (thus the validation set of the Pascal VOC 2007 is our selection set). Then we evaluate the selected 20 best concept detectors for all the 20 objects in VOC 2007 respectively. Note that for NUS-WIDE dataset, 4 object classes (bottle, table, sofa, tv) of Pascal VOC 2007 are not available in the 1000 provided tags. Hence, we could not learn the detectors of these classes from NUS-WIDE dataset.

Table~\ref{figure_detectionTable} displays the results obtained using our concept discovery algorithm on NUS-WIDE and SBU datasets and compares the state-of-the-art baselines with various kinds of supervision. CVPR'14~\cite{girshick2013rich} is the R-CNN detection framework, a fully supervised state-of-the-art method on Pascal VOC 2007. It uses the train set and validation set with bounding boxes to train the object detectors with deep features, then generates region proposal and deep feature for testing (we use the scores without fine-tuning). ICML'14~\cite{song2014learning} is the state-of-the-art method method for weakly supervised approaches on Pascal VOC 2007. It assumes that there are just image level labeling on the train set and validation set without bounding boxes to train the object detectors. It uses R-CNN framework to compute features on image windows to train the detectors and to generate region proposals and deep features for testing. ICCV'11~\cite{siva2011weakly} is another weakly supervised method using DPM. Since all these three methods only use the train set and validation set of Pascal VOC 2007 to train the detector, they are relevant to our method as ``upper bound" baselines. 

Another two most relevant comparison methods are the webly supervised method~\cite{2014webly} and video supervised method~\cite{siva2011weakly}. Webly supervised method uses items in Google N-grams as queries to collect images from image search engine for  training the detectors. So their training set of detector could be considered as the unlimited number of images from search engines. Video supervised method~\cite{siva2011weakly} trains detectors on manually selected videos without bounding boxes and shows results on 10 classes of Pascal VOC 2007. Since these two methods train detectors on other data source then test on Pascal VOC 2007, which is similar to our scenario, we consider them as direct comparison baselines. Our method outperforms these two methods with better AP on majority of the classes.

\section{Conclusion and Future Work}
\label{sec:Conclusion}
In this paper, we presented ConceptLearner, a max-margin hard instance learning approach to discover visual concepts from weakly labeled image collection. With more than 10,000 concept detectors learned from NUS-WIDE and SBU datasets, we apply the discovered concepts to concept recognition and detection. Based on the domain-selected supervision, we further quantitatively evaluate the learned concepts on  benchmarks for scene recognition and object detection, with promising results compared to other fully and weakly supervised methods. 

There are several possible extensions and applications for the discovered concepts. Firstly, since there are thousands of the concepts discovered, some concept detectors have overlaps. For example, as the predicted labels in the second example in Figure~\ref{figure_imagelevelannotation}(b), there are `market-fruit:NN',`market-local:AMOD',`market-a:DET',`market-vegetable:NN',`market-farmers:NN', and `fruit-market:PREP IN', which are redundant to describe the same image. Thus some bottom-up or top-down clustering methods could be used to merge the similar concept detectors or to merge the predicted labels for a query image. Besides, some measures could be introduced to characterize the properties of learned concepts, such as the visualness~\cite{jeong2012towards} and localizability~\cite{berg2010automatic}. Then the subset of concept detectors could be grouped and used in a specific image interpretation task. Meanwhile, in concept recognition and concept detection, since every concept is  detected independently, some spatial or co-occurrence constraints could be defined and used to filter out some outlier concepts detected in the same image, in the context of all the other detected concepts. Besides, with the grammatical structure integrated, the predicted phrases and tags could be further used to generate a full sentence description for the image.

{\small
\bibliographystyle{ieee}
\bibliography{ConceptLearner}
}

\end{document}